%% file: aaai-bestsubselect.tex
\documentclass[letterpaper]{article}

\input{commands}

\begin{document}
\title{Noisy Submodular Maximization via Adaptive Sampling \\with Applications to Crowdsourced Image Collection Summarization}
\author{
Adish Singla \\ ETH Zurich \\ adish.singla@inf.ethz.ch
\And
Sebastian Tschiatschek  \\ ETH Zurich \\ sebastian.tschiatschek@inf.ethz.ch
\And
Andreas Krause \\ ETH Zurich \\ krausea@ethz.ch 
}
\maketitle
\begin{abstract}
\begin{quote}
We address the problem of maximizing an unknown submodular function that can only be accessed via noisy evaluations. Our work is motivated by the task of summarizing content, e.g., image collections, by leveraging users' feedback in form of clicks or ratings. For summarization tasks with the goal of maximizing coverage and diversity, submodular set functions are a natural choice. When the underlying submodular function is unknown, users' feedback can provide noisy evaluations of the function that we seek to maximize. We provide a generic algorithm -- \submM{} -- for maximizing an unknown submodular function under cardinality constraints. This algorithm makes use of a novel exploration module -- \blbox{} -- that proposes good elements based on adaptively sampling noisy function evaluations. \blbox{} is able to accommodate different kinds of observation models such as value queries and pairwise comparisons.  We provide PAC-style guarantees on the quality and sampling cost of the solution obtained by \submM{}. 
We demonstrate the effectiveness of our approach in an interactive, crowdsourced image collection summarization application.
\end{quote}
\end{abstract}

\input{introduction}
\input{related}
\input{model}
\input{algoPart1_BestSubset}
\input{algoPart2_BestItem}
\input{experiments}
\input{conclusion}


\vspace{3mm}
\noindent {\footnotesize \textbf{Acknowledgments.}
We would like to thank Besmira Nushi for helpful discussions. This research is supported in part by SNSF grant 200021\_137971 and the Nano-Tera.ch program as part of the Opensense II project.}

\clearpage
\fontsize{9.5pt}{10.5pt}
\selectfont
\bibliographystyle{aaai}
\bibliography{aaai-bestsubselect}  

\clearpage
\input{appendix}

\end{document}

%% file: commands.tex
\usepackage{aaai16}
\usepackage{times}
\usepackage{helvet}
\usepackage{courier}
\usepackage{multirow}
\usepackage{xspace}
\usepackage{graphicx}
\usepackage{url}
\usepackage{enumitem}
\usepackage{color}
\usepackage{amsmath,amssymb,xfrac,amsthm,amsfonts}
\usepackage[tight]{subfigure}
\usepackage{verbatim, xfrac}
\usepackage[titletoc]{appendix}

\usepackage{subscript}
\setitemize{noitemsep,topsep=1pt,parsep=1pt,partopsep=1pt, leftmargin=12pt}
\setlist{nolistsep}

\newtheorem{theorem}{Theorem}

\makeatletter

\newcommand{\Rmnum}[1]{\expandafter\@slowromancap\romannumeral #1@}
\makeatother
\usepackage{caption}

\usepackage[lined, boxed, ruled, commentsnumbered, noend]{algorithm2e}

\usepackage{algpseudocode}

\newcommand{\citet}[1]{\citeauthor{#1} (\citeyear{#1})}

\DeclareMathOperator*{\argmax}{arg\,max}

\newcommand{\actionset}{\ensuremath{\mathcal{V}}}
\newcommand{\submM}{\textsc{ExpGreedy}\xspace}

\newcommand{\blbox}{\textsc{TopX}\xspace}

\newcommand{\nrItems}{\ensuremath{k}}

\newcommand{\greedy}{\textsc{Greedy}\xspace}
\newcommand{\random}{\textsc{Random}\xspace}

\newcommand{\uniform}{\textsc{Uniform}\xspace}



%% file: introduction.tex

\section{Introduction}\label{sec.introduction}

\begin{figure*}[!t]
\centering
   \subfigure[Image collection to be summarized]{
     \includegraphics[width=0.44\textwidth]{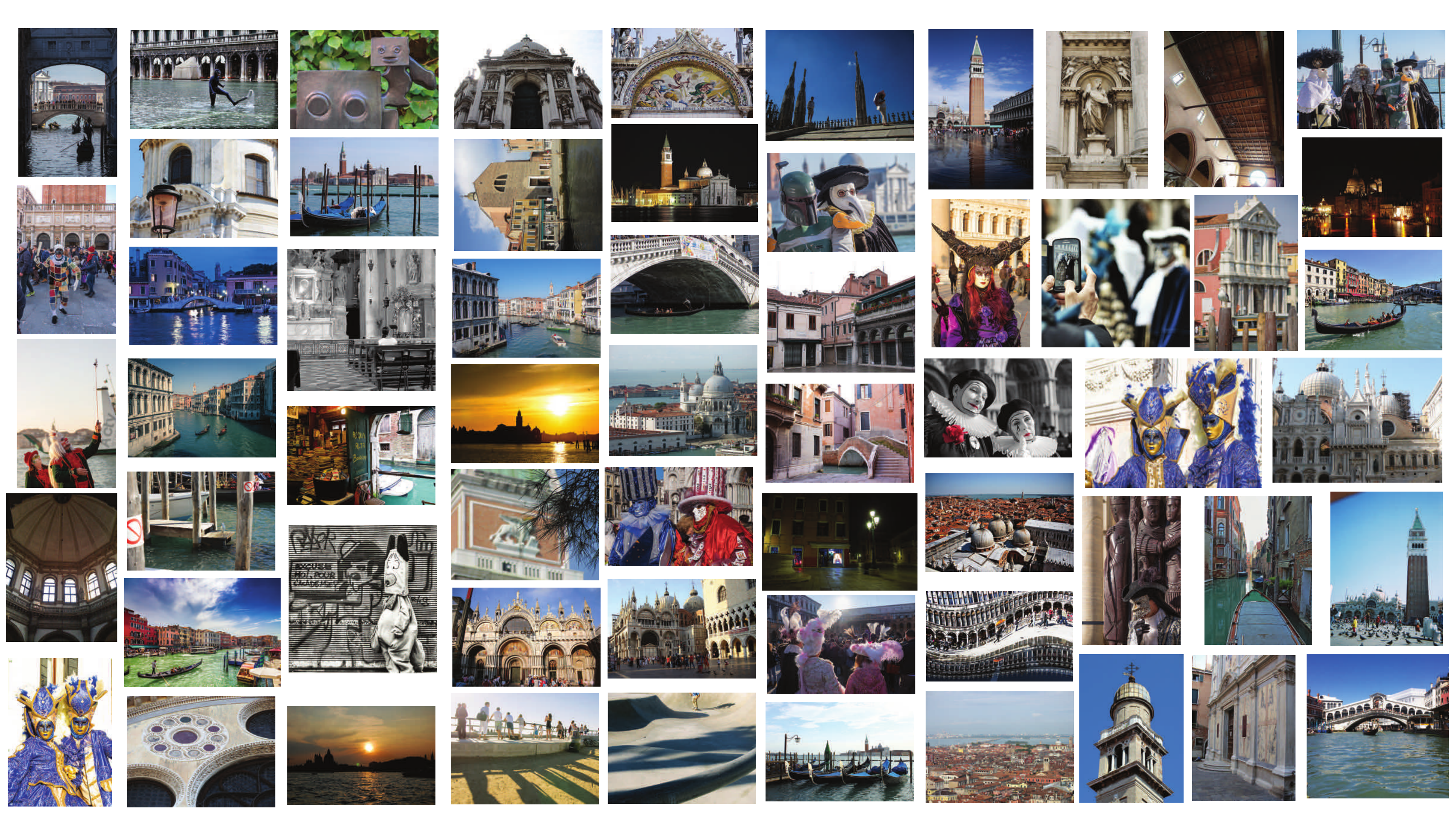}
     \label{fig.results.collection}
   }
   \subfigure[Results for three summarization tasks]{
     \includegraphics[width=0.29\textwidth]{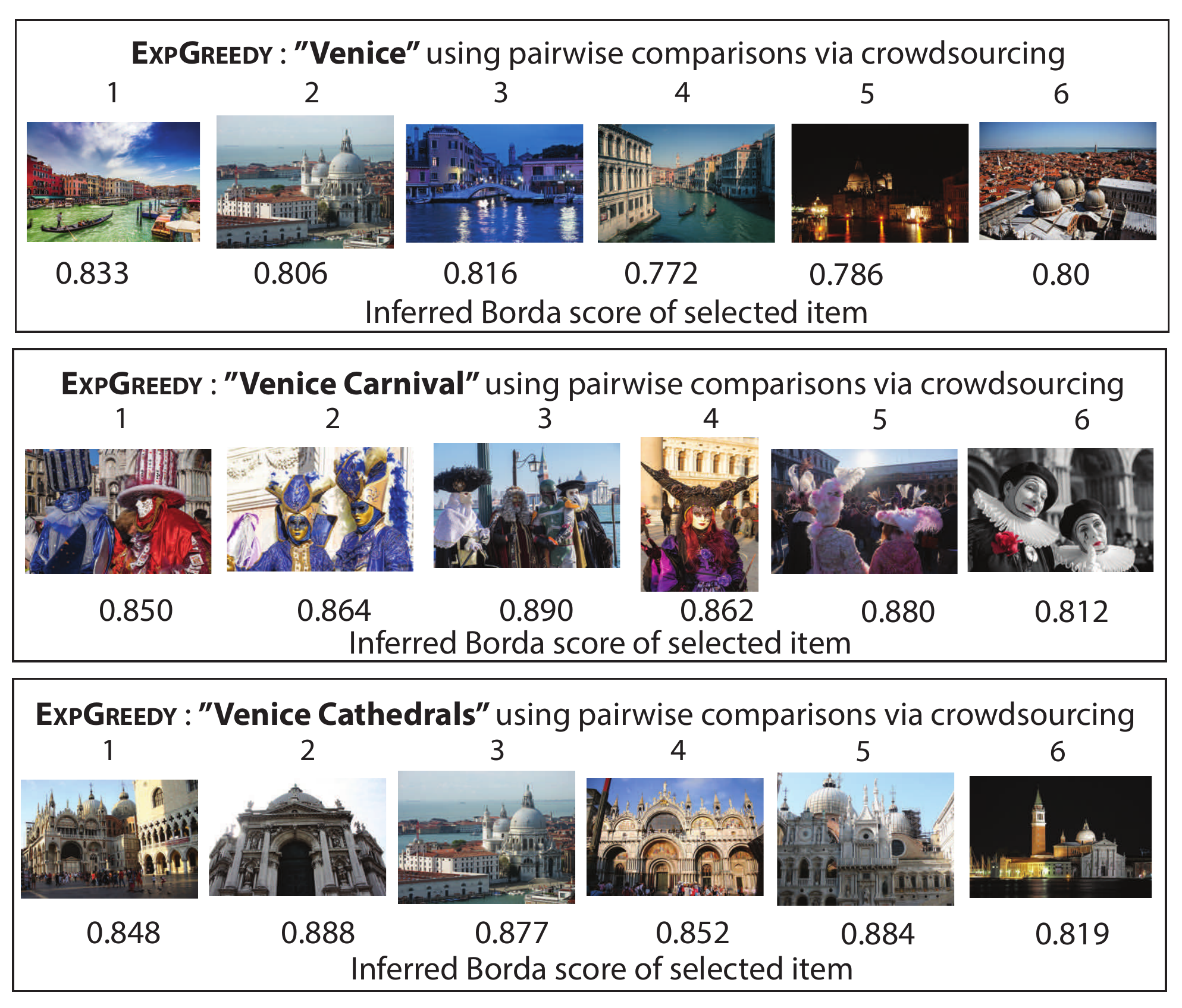}
    \label{fig.results.crowdsummary}
   }
\caption{\footnotesize {\bf (a)} Image collection $\actionset$ to be summarized; {\bf (b)} Summaries obtained using pairwise comparisons via crowdsourcing for the themes $(i)$ Venice,   $(ii)$ Venice Carnival,  and $(iii)$ Venice Cathedrals. Images are numbered from $1$ to $6$ in the order of selection. }
\label{fig.results.main}
\end{figure*}


Many applications involve the selection of a subset of items, e.g., summarization of content on the web. 
Typically, the task is to select a subset of items of limited cardinality with the goal of maximizing their utility. This utility is often measured via properties like diversity, information, relevance or coverage. Submodular set functions naturally capture the fore-mentioned notions of utility. Intuitively, submodular functions are set functions that satisfy a natural diminishing returns property. They have been widely used in diverse applications, including content summarization and recommendations, sensor placement, viral marketing, and numerous machine learning and computer vision tasks~\cite{2008_beyond_convexity,2015_submodularity_in_machine_learning}.

{\bf Summarization of image collections.}
One of the motivating applications, which is the subject of our experimental evaluation, is summarization of image collections. Given a collection of images, say, from \emph{Venice} on Flickr, we would like to select a small subset of these images that summarizes the theme \emph{Cathedrals in Venice}, cf., Figure~\ref{fig.results.main}. We cast this summarization task as the problem of maximizing a submodular set function $f$ under cardinality constraints. 

{\bf Submodular maximization.}
Usually, it is assumed that $f$ is known, i.e., $f$ can be evaluated exactly by querying an oracle. In this case, a greedy algorithm is typically used for maximization. The greedy algorithm builds up a set of items by picking those of highest marginal utility in every iteration, given the items selected that far. 
Despite its greedy nature, this algorithm provides the best constant factor approximation to the optimal solution computable in polynomial time~\cite{1978-_nemhauser_submodular-max}.

{\bf Maximization under noise.} However, in many realistic applications, the function $f$ is {\em not known} and can only be evaluated up to (additive) noise.
For instance, for the image summarization task, (repeatedly) querying users' feedback in form of clicks or ratings on the individual images or image-sets can provide such noisy evaluations. There are other settings for which marginal gains are hard to compute exactly, e.g., computing marginal gains of nodes in viral marketing applications~\cite{kempe2003maximizing} or conditional information gains in feature selection tasks~\cite{krause05near}.
In such cases, one can apply a naive \emph{uniform} sampling approach to estimate all marginal gains up to some error $\epsilon$ and apply the standard greedy algorithm. While simple, this uniform sampling approach could have high sample complexity, rendering it impractical for real-world applications. In this work, we propose to use \emph{adaptive} sampling strategies to reduce sample complexity while maintaining high quality of the solutions.  

{\bf Our approach for adaptive sampling.}
The first key insight for efficient adaptive sampling is that we need to estimate the marginal gains only up to the necessary confidence to decide for the best item to select next. 
However, if the difference in the marginal gains of the best item and the second best item is small, this approach also suffers from high sample complexity. 
To overcome this, we exploit our second key insight that instead of focusing on selecting the single best item, it is sufficient to select an item from a small subset of {\em any} size $l \in \{1, \ldots, \nrItems\}$ of high quality items, where $\nrItems$ is the cardinality constraint. 
Based on these insights, we propose a novel exploration module -- \blbox{} -- that proposes good elements by adaptively sampling noisy function evaluations.

{\bf Our contributions.}
Our main contributions are:
\begin{itemize}
 \item We provide a greedy algorithm \submM{} for submodular maximization of a function via noisy evaluations. The core part of this algorithm, the exploration module \blbox{}, is invoked at every iteration and implements a novel adaptive sampling scheme for efficiently selecting a small set of items with high marginal utilities.
 \item Our theoretical analysis and experimental evaluation provide insights of how to trade-off the quality of subsets selected by \submM{} against the number of evaluation queries performed by \blbox{}. 
 \item We demonstrate the applicability of our algorithms in a real-world application of crowdsourcing the summarization of image collections via eliciting crowd preferences based on (noisy) pairwise comparisons, cf., Figure~\ref{fig.results.main}.
\end{itemize}

%% file: related.tex
\vspace{-1.5mm}
\section{Related Work}\label{sec.related}
\vspace{-1mm}

{\bf Submodular function maximization (offline).} 
Submodular set functions $f(S)$ arise in many applications and, therefore, their optimization has been studied extensively. For example, a celebrated result of~\citet{1978-_nemhauser_submodular-max} shows that non-negative monotone submodular functions under cardinality constraints can be maximized up to a constant factor of $(1-1/e)$ by a simple greedy algorithm. Submodular maximization has furthermore been studied for a variety of different constraints on $S$, e.g., matroid constraints or graph constraints~\cite{2006_near_optimal_sensor_placement,2009_nonmyopic_adaptive_information_planning}, and in different settings, e.g., distributed optimization~\cite{2013_distributed_submodular_maximization}.  When the function $f$ can only be evaluated up to (additive) noise, a naive \emph{uniform} sampling approach has been employed to estimate all the marginal gains \cite{kempe2003maximizing,krause05near}, an approach that could have high sampling complexity.

{\bf Learning submodular functions.} 
One could approach the maximization of an unknown submodular function by first learning the function of interest and subsequently optimizing it. \citet{2014_image_summarization}~present an approach for learning linear mixtures of known submodular component functions for image collection summarization. Our work is complimentary to that approach wherein we directly target the subset selection problem without learning the underlying function. In general, learning submodular functions from data is a difficult task --- \citet{2011_learning_submodular_functions}~provide several negative results in a PAC-style setting. 

{\bf Submodular function maximization (online).} 
Our work is also related to online submodular maximization with (opaque) bandit feedback. \citet{2008_online_submodular_max}~present approaches for maximizing a sequence of submodular functions in an online setting. Their adversarial setting forces them to use conservative algorithms with slow convergence. \citet{2011_submodular_bandits}~study a more restricted setting, where the objective is an (unknown) linear combination of known submodular functions, under stochastic noise. While related in spirit, these approaches aim to minimize cumulative regret. In contrast, we aim to identify a single good solution performing as few queries as possible --- the above mentioned results do not apply to our setting.

{\bf Best identification (pure exploration bandits).}
In exploratory bandits, the learner first explores a set of actions under time / budget constraints and then exploits the gathered information by choosing the estimated best action ($top^1$ identification problem)~\cite{2006-jlmr_explore-1,2009-alt_pure-exploration}.  Beyond best individual actions, \citet{2014-icml_pac-mab} design an $(\epsilon, \delta)$-PAC algorithm for the $top^m$ identification problem where the goal is to return a subset of size $m$ whose aggregate utility is within $\epsilon$ compared to the aggregate utility of the $m$ best actions. \citet{chen2014combinatorial} generalize the problem by considering combinatorial constraints on the subsets that can be selected, e.g., subsets must be size $m$, represent matchings, etc. They present general learning algorithms for all decision classes that admit offline maximization oracles. Dueling bandits are variants of the bandit problem where feedback is limited to relative preferences between pairs of actions. The best-identification problem is studied in this weaker information model for various notions of ranking models (e.g., {\em Borda} winner), cf.,~\citet{busa2014survey}. However, in contrast to our work, the reward functions considered by~\citet{chen2014combinatorial} and other existing algorithms are modular, thus limiting the applicability of these algorithms. 
Our work is also related to contemporary work by \citet{singer}, who treat submodular maximization under noise. While their algorithms apply to persistent noise, their technique is computationally demanding, and does not enable one to use noisy preference queries.

%% file: model.tex
\section{Problem Statement}\label{sec.model}

{\bf Utility model.}  
Let $\actionset=\{1, 2,\ldots, N\}$ be a set of $N$ items.  We assume a utility function $f\colon 2^\actionset \rightarrow \mathbb{R}$ over subsets of $\actionset$. Given a set of items $S \subseteq \actionset$, the utility of this set is $f(S)$. Furthermore, we assume that $f$ is \emph{non-negative}, \emph{monotone} and \emph{submodular}.  Monotone set functions satisfy $f(S) \leq f(S')$ for all $S \subseteq S' \subseteq \actionset$; and submodular functions satisfy the following diminishing returns condition: for all $S \subseteq S' \subseteq \actionset \setminus \{a\}$, it holds that $f(S \cup \{a\}) - f(S) \geq f(S' \cup\{a\})-f(S')$. These conditions are satisfied by many realistic, complex utility functions \cite{krause2011submodularity,2012-survey_krause_submodular}. Concretely, in our image collection summarization example, $\actionset$ is a collection of images, and $f$ is a function that assigns every summary $S\subseteq\actionset$ a score, preferring relevant and diverse summaries.  

{\bf Observation model.} 
In classical submodular optimization, $f$ is assumed to be known, i.e., $f$ can be evaluated exactly by an {\em oracle}. In contrast, we only assume that noisy evaluations of $f$ can be obtained. For instance, in our summarization example, one way to evaluate $f$ is to query users to rate a summary, or to elicit user preferences via pairwise comparisons of different summaries. In the following, we formally describe these two types of queries in more detail:

\emph{(1) Value queries.} In this variant, we query the value for $f(a | S) = f(\{a\} \cup S) - f(S)$ for some $S \subseteq \actionset$ and $a \in \actionset$. We model the noisy evaluation or response to this query by a random variable $X_{a|S}$ with unknown sub-Gaussian distribution. We assume that $X_{a|S}$ has mean $f(a | S)$ and that repeated queries for $f(a|S)$ return samples drawn i.i.d.\ from the unknown distribution.
 
\emph{(2) Preference queries.} Let $a,b \in \actionset$ be two items and $S \subseteq \actionset$. The preference query aims at determining  whether an item $a$ is preferred over item $b$ in the context of $S$ (i.e., item $a$ has larger marginal utility than another item $b$). We model the noisy response of this pairwise comparison by the random variable $X_{a > b|S}$ that takes values in $\{0,1\}$. We assume that $X_{a > b|S}$ follows some unknown distribution and satisfies the following two properties: (i) $X_{a>b|S}$ has mean larger than $0.5$ iff $f(a | S) > f(b |S)$; (ii) the mapping from utilities to probabilities is monotone in the sense, that if given some set $S$, and given that the gaps in utilities satisfy $f(a|S)-f(b|S) \geq f(a'|S)-f(b'|S)$ for items $a,b,a',b' \in \actionset$, then the mean of $X_{a>b|S}$ is greater or equal to the mean of $X_{a'>b'|S}$. For instance, the distribution induced by the commonly used Bradley-Terry-Luce preference model~\cite{1952_rank_analysis,1959_individual_choice_behavior} satisfies these conditions. 
We again assume that repeated queries return samples drawn i.i.d.\ from the unknown distribution. 

Value queries are natural, if $f$ is approximated via stochastic simulations (e.g., as in \citet{kempe2003maximizing}). 
On the other hand, preference queries may be a more natural way to learn what is relevant / interesting to users compared to asking them to assign numerical scores, which are difficult to calibrate.

{\bf Objective.}
Our goal is to select a set of the items $S\subseteq \actionset$ with $|S| \leq \nrItems$ that maximizes the utility $f(S)$. The optimal solution to this problem is given by 
\begin{equation}
  S^{opt} = \argmax_{S \subseteq \actionset, |S| \leq \nrItems} f(S).
  \label{eq:optset}
\end{equation}
Note that obtaining optimal solutions to problem~\eqref{eq:optset} is intractable~\cite{1998-_feige_threshold-of-ln-n}. However, a greedy optimization scheme based on the marginal utilities of the items can provide a solution $S^{greedy}$ such that $f(S^{greedy}) \geq (1 - \frac{1}{e}) \cdot f(S^{opt})$, i.e., a solution that is within a constant factor of the optimal solution can be efficiently determined.

In our setting, we can only evaluate the unknown utility function $f$ via noisy queries and thus cannot hope to achieve the same guarantees.  The key idea is that in a stochastic setting, our algorithms can make repeated queries and aggregate noisy evaluations to obtain sufficiently accurate estimates of the marginal gains of items.
We study our proposed algorithms in a PAC setting, i.e., we aim to design algorithms that,  given positive constants $(\epsilon, \delta)$, determine a set $S$that is $\epsilon$-competitive relative to a reference solution with probability of at least $1 - \delta$.
One natural baseline is a constant factor approximation to the optimal solution, i.e., we aim to determine a set $S$ such that with probability at least $1-\delta$,
\begin{equation}
  f(S) \geq (1 - \frac{1}{e}) \cdot f(S^{opt}) - \epsilon.
  \label{optsetepsilon2}
\end{equation}
Our objective is to achieve the desired $(\epsilon, \delta)$-PAC guarantee while minimizing sample complexity (i.e., the number of  evaluation queries performed). 

%% file: algoPart1_BestSubset.tex
\section{Submodular Maximization Under Noise}\label{sec.algoPart1}

We now present our algorithm \submM{} for maximizing submodular functions under noise. Intuitively, it aims to mimic the greedy algorithms in noise-free settings, ensuring that it selects a good element in each iteration. Since we cannot evaluate the marginal gains exactly, we must experiment with different items, and use statistical inference to select items of high value. This experimentation, the core part of our algorithm, is implemented via a novel exploration module called \blbox{}. 

The algorithm \submM{}, cf., Algorithm~\ref{alg.submset}, iteratively builds up a set $S \subseteq \actionset$ by invoking $\blbox(\epsilon', \delta', k', S)$ at every iteration to select the next item. \blbox{} returns candidate items that could potentially be included in $S$ to maximize its utility. 
The simplest adaptive sampling strategy that \blbox{} could implement is to estimate the marginal gains up to the necessary confidence to decide for the next best item to select ($top^1$ identification problem). 
However, if the difference in the marginal gains of the best and the second best item is small, this approach could suffer from high sample complexity. Extending ideas from the randomized greedy algorithm \cite{2014_submodular_maximization_with_cardinality_constraints}, we show in Theorem~\ref{thm2} that in every iteration, instead of focusing on $top^1$, it is sufficient for \submM{} to select an item from a small subset of
items with high utility. This corresponds to the following two conditions on \blbox{}:
 
 



\begin{enumerate}
  \item \blbox{} returns a subset $A \neq \emptyset$ of size {\em at most} $k'$.
  \item With probability at least $1-\delta'$, the items in $A$ satisfy 
    \begin{equation}
      \frac{1}{|A|} \sum_{a \in A} f(a|S) \geq \mathop{\max_{B \subseteq \actionset}}_{|B| = |A|} \left[ \frac{1}{|B|} \sum_{b \in B} f(b|S) \right]  - \epsilon'. \label{cond.subsetop}
    \end{equation}
\end{enumerate}

\begin{algorithm}[t!]
  \nl {\bf Input}: {Ground set $\actionset$; No.~of items to pick $\nrItems$; $\epsilon, \delta>0$; }\\
\nl {\bf Output}: {Set of items $S \subseteq \actionset\colon |S| \leq \nrItems$, such that $S$ is $\epsilon$-competitive with probability at least $(1 - \delta)$;} \\
\nl {\bf Initialize}: $S = \emptyset$\\
  \ForEach {$j = 1, \ldots, \nrItems$}{
	\nl $A = \blbox(\epsilon', \delta', k', S)$\\
	\nl Sample $s$ uniformly at random from $A$\\
    \nl $S = S \cup \{s\}$\\
  }
  \nl {\bf return} $S$\\
  \caption{\submM{}}
  \label{alg.submset} 
\end{algorithm}

At any iteration, satisfying these two conditions is equivalent to solving a $top^l$ identification problem for $l = |A|$ with PAC-parameters $(\epsilon', \delta')$. \blbox{} essentially returns a set of size $l$ containing items of largest marginal utilities given $S$. Let us consider two special cases, (i) $l=1$ and (ii) $l=\nrItems$ for all $j \in \{1, \ldots, \nrItems\}$ iterations. Then, in the noise free setting, \submM{} for case (i) mimics the classical greedy algorithm~\cite{1978-_nemhauser_submodular-max} and for case (ii) the randomized greedy algorithm~\cite{2014_submodular_maximization_with_cardinality_constraints}, respectively.  As discussed in the next section, the sample complexity of these two cases can be very high. The key insight we use in \submM{} is that if we could efficiently solve the $top^l$ problem for {\em any} size $l \in \{1, \ldots, \nrItems\}$ (i.e., $|A|$ is neither necessarily $1$ or $\nrItems$), then the solution to the submodular maximization problem is guaranteed to be of high quality. This is summarized in the following theorem:

%
%
\begin{theorem} \label{thm2} Let $\epsilon > 0, \delta \in (0,1)$. Using $\epsilon' = \tfrac{\epsilon}{\nrItems}, \delta' = \tfrac{\delta}{\nrItems}$ and $k' = \nrItems$ for invoking \blbox, Algorithm~\ref{alg.submset} returns a set $S$ that satisfies $\mathbb{E}[f(S)] \geq (1 - \frac{1}{e}) \cdot f(S^{opt}) - \epsilon$ with probability at least $1-\delta$. For the case that $k'=1$, the guarantee is $f(S) \geq (1 - \frac{1}{e}) \cdot f(S^{opt}) - \epsilon$ with probability at least $1-\delta$. 
\end{theorem}
The proof is provided in Appendix~A of the extended version of paper \cite{singla16noisy-longer}.

As it turns out, solutions of the greedy algorithm in the noiseless setting $S^{greedy}$ often have utility larger than $(1 - \frac{1}{e}) \cdot f(S^{opt})$.
Therefore, we also seek algorithms that with probability at least $1-\delta$ identify solutions satisfying 
\begin{align}
  f(S) \geq f(S^{greedy})  - \epsilon.
  \label{optsetepsilon1}
\end{align}
This can be achieved according to the following theorem:
\begin{theorem} \label{thm1} Let $\delta \in (0,1)$. Using $\epsilon' = 0, \delta' = \tfrac{\delta}{\nrItems}$ and $k'=1$ for invoking \blbox, Algorithm~\ref{alg.submset} returns a set $S$ that satisfies $f(S) = f(S^{greedy})$ with probability at least $1-\delta$.
\end{theorem}
If we set $k'=1$ and $\epsilon=0$, then condition \eqref{cond.subsetop} is actually equivalent to requiring that \blbox, with high probability, identifies the element with largest marginal utility. The proof follows by application of the union bound. This theorem ensures that if we can construct a corresponding exploration module, we can successfully compete with the greedy algorithm that has access to $f$. However, this can be prohibitively expensive in terms of the required number of queries performed by \blbox, cf.,   Appendix~B of the extended version of this paper \cite{singla16noisy-longer}.

%% file: algoPart2_BestItem.tex
\section{Exploration Module \blbox{}}\label{sec.algoPart2}

\begin{figure}[t!]
  \centering
     \includegraphics[trim = {20mm 0mm 20mm 0mm}, clip=true, width=1\linewidth]{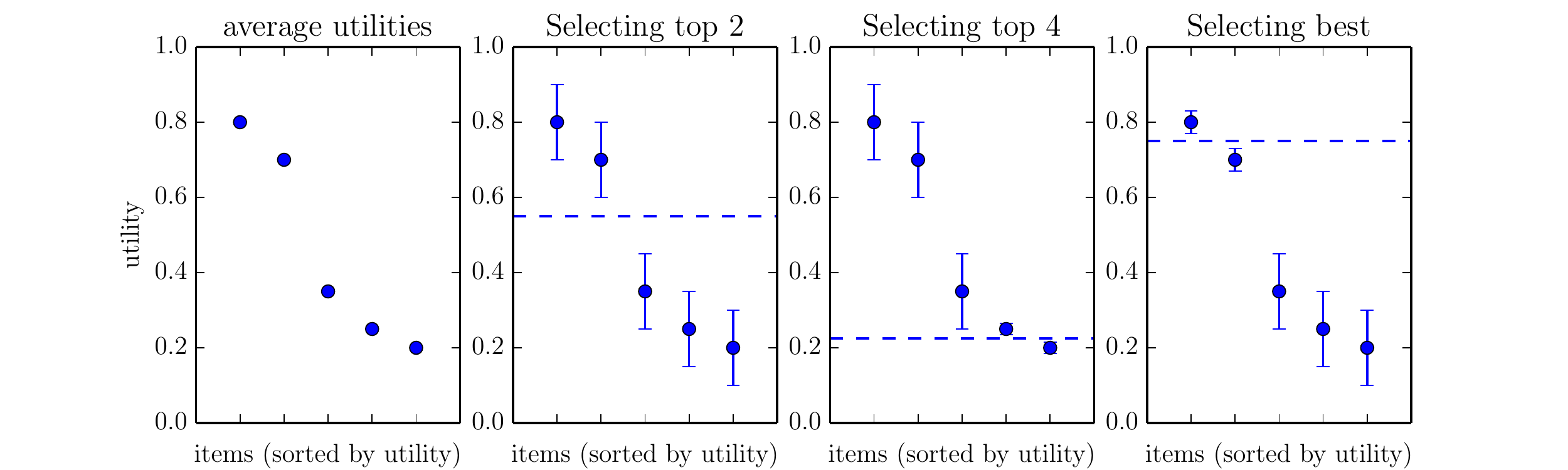}
  \caption{An example to illustrate the idea of $top^l$ item selection with $N=5$ items and parameter $k'=4$. While both the $top^1$ identification ($l=1$) and the $top^4$ identification ($l=k'$) have high sample complexity, the $top^2$ identification ($l=2 \in \{1, \ldots, k'\}$) is relatively easy.}
  \label{fig.sel}
\end{figure}

In this section, we describe the design of our exploration module \blbox{} used by \submM{}. 

\subsection{\blbox{} with Value Queries}
We begin with the observation that for fixed $l \in \{1, \ldots, k'\}$ (for instance $l=1$ or $l=k'$), an exploration module \blbox{} satisfying condition~\eqref{cond.subsetop} can be implemented by solving a $top^l$ identification problem. 
This can be seen as follows. 

Given input parameter $S$, for each $a\in\actionset\setminus S$ define its value $v_a=f(a|S)$, and define $v_a=0$ for each $a\in S$. Then, the value of any set $A$ (in the context of $S$) is $v(A) := \sum_{a\in A} v_a$, which is a modular (additive) set function. Thus, in this setting (fixed $l$), meeting condition \eqref{cond.subsetop} requires identifying a set $A$ maximizing a {\em modular} set function under cardinality constraints from noisy queries.  This corresponds to the $top^l$ best-arm identification problem. In particular, \citet{chen2014combinatorial}~proposed an algorithm -- CLUCB --  for identifying a best subset for modular functions under combinatorial constraints. At a high level, CLUCB maintains upper and lower confidence bounds on the item values, and adaptively samples noisy function evaluations until the pessimistic estimate (lower confidence bound) for the current best $top^l$ subset exceeds the optimistic estimate (upper confidence bound) of any other subset of size $l$.
The worst case sample complexity of this problem is characterized by the gap between the values of $l$-th item and $(l+1)$-th item (with items indexed in descending order according to the values $v_a$).

As mentioned, the sample complexity of the $top^l$ problem can vary by orders of magnitude for different values of $l$, cf., Figure~\ref{fig.sel}.  Unfortunately, we do not know the value of $l$ with the lowest sample complexity in advance. However, we can modify CLUCB to jointly estimate the marginal gains and solve the $top^l$ problem with lowest sample complexity.


The proposed algorithm implementing this idea is presented in Algorithm~\ref{alg.modmax}. It maintains confidence intervals for marginal item values, with the confidence radius of an item $i$ computed as $rad_t(i) = R \sqrt{2 \log \left( \frac{4Nt^3}{\delta'}\right) / T_t(i)},$ where the corresponding random variables $X_{i|S}$ are assumed to have an $R$-sub-Gaussian tail and $T_t(i)$ is the number of observations made for item $i$ by time $t$. If the exact value of $R$ is not known, it can be upper bounded by the range (as long as the variables have bounded centered range). Upon termination, the algorithm returns a set $A$ that satisfies condition~\eqref{cond.subsetop}. Extending results from~\citet{chen2014combinatorial}, we can bound the sample complexity of our algorithm as follows. Consider the gaps  $\Delta_{l}=f(\pi(l)|S) - f(\pi(l+1)|S)$, where
 $\pi\colon \actionset \rightarrow  \{ 1, \ldots, N\} $ is a permutation of the items such that $f(\pi(1)|S) \geq f(\pi(2)|S) \geq \ldots \geq f(\pi(N)|S)$. 
For every {\em fixed} $l$, the sample complexity of identifying a set of top $l$ items is characterized by this gap $\Delta_{l}$. 
The reason is that if $\Delta_l$ is small, many samples are needed to ensure that the confidence bounds $rad_t$ are small enough to distinguish the top $l$ elements from the runner-up. Our key insight is that we can be adaptive to the largest $\Delta_l$, for $l\in \{1,\dots,k'\}$. That is, as long as there is {\em some} value of $l$ with large $\Delta_l$, we will be able to enjoy low sample complexity (cf., Figure~\ref{fig.sel}):
%
\begin{theorem} \label{thm.sample.complexity} Given $\epsilon' > 0, \delta' \in (0,1)$, $S \subseteq \actionset$ and $k'$, Algorithm~\ref{alg.modmax} returns a set $A \subseteq \actionset, |A| \leq k'$ that with probability at least $1-\delta'$ satisfies condition \eqref{cond.subsetop} using at most
\begin{equation*}
  T \leq \mathcal{O}\left( k' \min_{l=1, \ldots, k'} \left[ R^2   H_{(l,\epsilon')}  \log \left(\frac{R^2}{\delta'}  H_{(l,\epsilon')}  \right) \right] \right) 
\end{equation*}
samples, where $H_{(l,\epsilon')} = N \min \{ \tfrac{4}{\Delta_{l}^{2}}, \tfrac{1}{\epsilon'^{2}} \}$. 
\end{theorem}
%
\noindent A proof sketch is given in Appendix~C of the extended version of this paper~\cite{singla16noisy-longer}.

\begin{algorithm}[t!]
  \nl {\bf Input}: {Ground set $\actionset$; Set $S$; integer $k'$; $\epsilon', \delta'>0$;} \\
\nl {\bf Output}: {Set of items $A \subseteq \actionset: |A| \leq  k'$, such that $A$ satisfies~\eqref{cond.subsetop} with probability at least $1 - \delta'$;} \\
\nl {\bf Initialize}: For all $i=1,\ldots,|\actionset|$: observe $X_{i|S}$ and set $v_i$ to that value, set $T_1(i)=1$; \\
  \For {$t = 1, \ldots$}{
    \nl Compute confidence radius $rad_t(i)$ for all $i$;\\
    \nl Initialize list of items to query $Q = []$; \\
    \For {$l \in 1, \ldots, k'$}{
	  \nl $M_t = \argmax_{B \subseteq \actionset, |B| = l} \sum_{i \in B} v_{i};$\\
  	  \nl \ForEach{$i = 1, \ldots, |\actionset|$}{
	    \nl If $i \in M_t$ set $\widetilde{v}_i = v_i - rad_t(i)$, otherwise set $\widetilde{v}_i = \overline{v}_i + rad_t(i)$
	    }
	    \nl $\widetilde{M}_t = \argmax_{B \subseteq \actionset, |B| = l} \sum_{i \in B} \widetilde{v}_{i};$\\
	    \nl \If{$[\sum_{i \in \widetilde{M}_t} \widetilde{v}_i - \sum_{i \in M_t} \widetilde{v}_i] \leq l \cdot \epsilon'$}{
	       \nl Set $A=M_t$ and {\bf return} $A$
	   }
	\nl Set $q = \argmax_{i \in (M \setminus \widetilde{M}) \cup (\widetilde{M} \setminus M)} rad_t(i)$;\\
	\nl Update list of items to query: $Q = Q.append(q)$ 
	}
	\ForEach {$q \in Q$}{
	  \nl Query and observe output $X_{q|S}$;\\
      \nl Update empirical means $v_{t+1}$ using the output;\\
	  \nl Update observation counts $T_{t+1}(q) = T_{t}(q) + 1$ and $T_{t+1}(j) = T_{t}(j)$ for all $j \neq q$;\\
	}
  }
  
 \caption{\blbox{}}  
  \label{alg.modmax} 
\end{algorithm}

\subsection{\blbox{} with Preference Queries}
We now show how noisy preference queries can be used. As introduced previously, we assume that there exists an underlying preference model (unknown to the algorithm) that induces probabilities $P_{i>j|S}$ for item $i$ to be preferred over item $j$ given the values $f(i|S)$ and  $f(j|S)$.  
In this work, we focus on identifying the {\em Borda winner}, i.e., the item $i$ maximizing the Borda score $P(i|S)$, formally given as $\tfrac{1}{(N-1)} \cdot {\sum_{j \in \actionset \setminus \{i\}} P_{i>j|S}}$. The Borda score measures the probability that item $i$ is preferred to another item chosen uniformly at random. Furthermore, in our model where we assume that an increasing gap in the utilities leads to monotonic increase in the induced probabilities, it holds that the top $l$ items in terms of marginal gains are the top $l$ items in terms of Borda scores.  
%
We now make use of a result called \emph{Borda reduction}~\cite{2015_sparse_dueling_bandits,busa2014survey}, a technique that allows us to reduce preference queries to value queries, and to consequently invoke Algorithm~\ref{alg.modmax}  with small modifications.

{\bf Defining values $v_i$ via Borda score}.  For each item $i \in \actionset$, Algorithm~\ref{alg.modmax} (step~$3,15$) tracks and updates mean estimates of the values $v_i$. For preference queries, these values are replaced with the Borda scores. The sample complexity in Theorem~\ref{thm.sample.complexity} is then given in terms of these Borda scores. For instance, for the Bradley-Terry-Luce preference model~\cite{1952_rank_analysis,1959_individual_choice_behavior}, the Borda score for item $i$ is given by $\tfrac{1}{(N-1)} \cdot {\sum_{j \in \actionset \setminus \{i\}} \tfrac{1}{1 + \exp{(-\beta(f(i|S) - f(j|S)))}}}$. Here, $\beta$ captures the problem difficulty: $\beta \to \infty$ corresponds to the case of noise-free responses, and $\beta \to 0$ corresponds to uniformly random binary responses. The effect of $\beta$ is further illustrated in the synthetic experiments.

{\bf Incorporating noisy responses.}
Observing the value for item $i$ in the preference query model  corresponds to pairing  $i$ with an item in the set  $\actionset \setminus \{i\}$ selected uniformly at random. The observed preference response provides an unbiased estimate of the Borda score for item $i$. More generally, we can pick a small set $Z_i$ of fixed size $\tau$ selected uniformly at random with replacement from $\actionset \setminus \{i\}$, and compare $i$ against each member of $Z_i$. Then, the observed Borda score for item $i$ is calculated as $\tfrac{1}{\tau} \cdot \sum_{j \in Z_i} X_{i > j|S}$. Here, $\tau$ is a parameter of the algorithm. One can observe that the cost of one preference query is $\tau$ times the cost of one value query. The effect of $\tau$ will be further illustrated in the experiments.

The subtle point of terminating Algorithm~\ref{alg.modmax} (step 10) is discussed in Appendix~D of the extended version of this paper \cite{singla16noisy-longer}.

%% file: experiments.tex
\section{Experimental Evaluation}\label{sec.experiments.syn}
We now report on the results of our synthetic experiments.

\begin{figure*}[!t]
\centering
   \subfigure[Sample complexity with varying noise]{
     \includegraphics[width=0.29\textwidth]{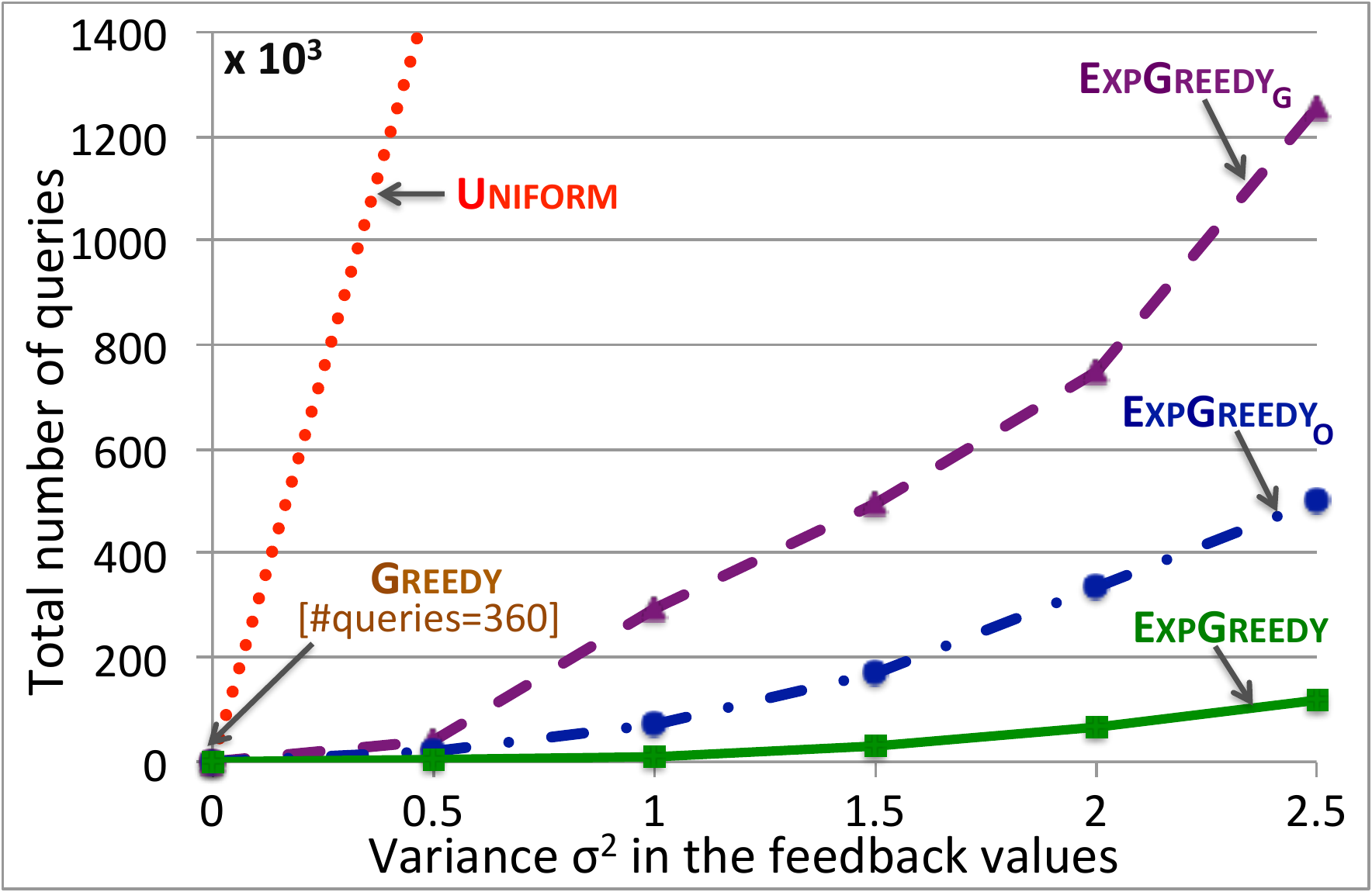}
    \label{fig.results.sample-complexity}
   }
   \subfigure[Distribution of queries across items]{
     \includegraphics[width=0.29\textwidth]{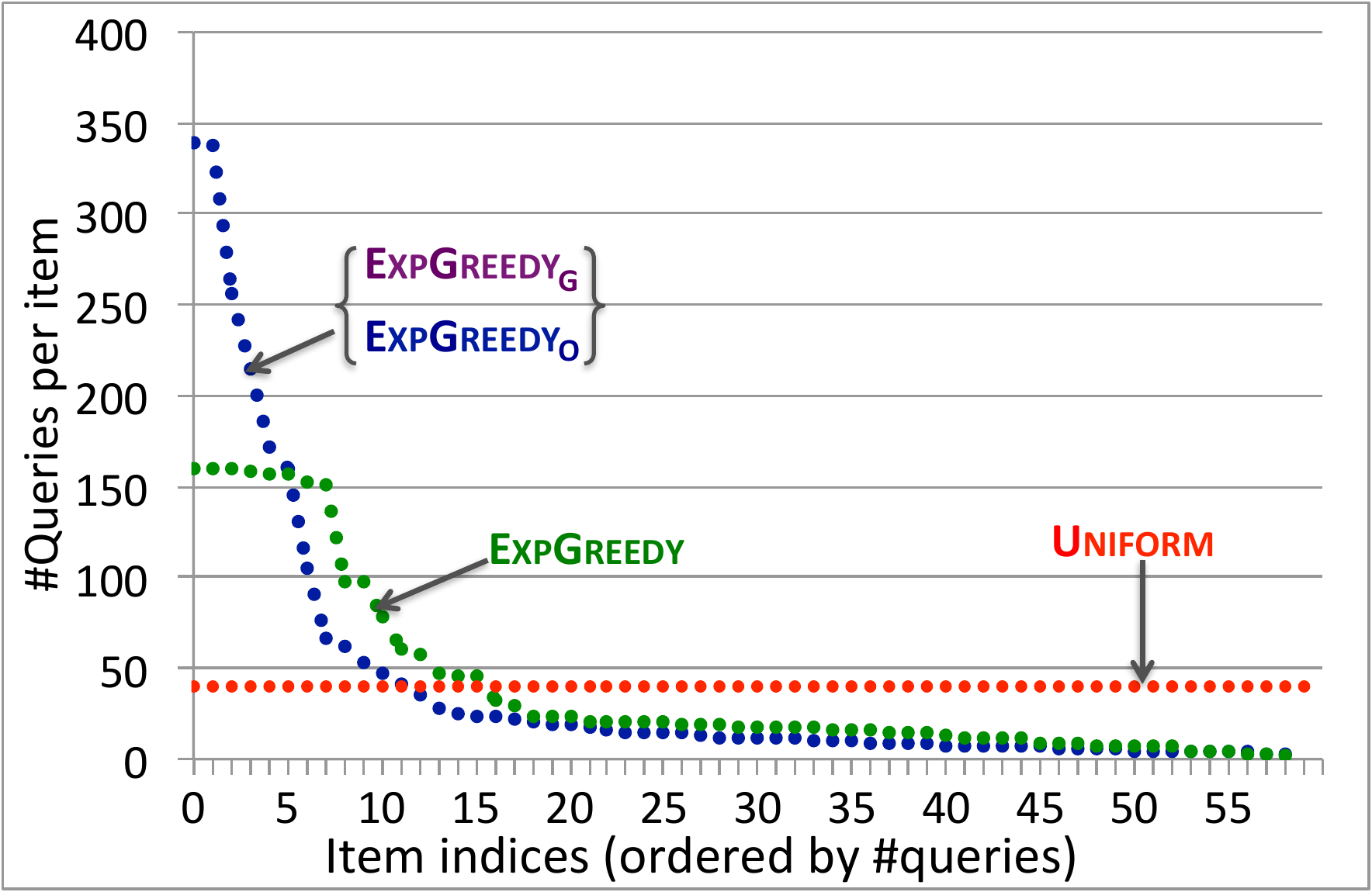}
     \label{fig.results.budget-allocation}
   }   
   \subfigure[Execution instance showing $top^l$ size]{
     \includegraphics[width=0.29\textwidth]{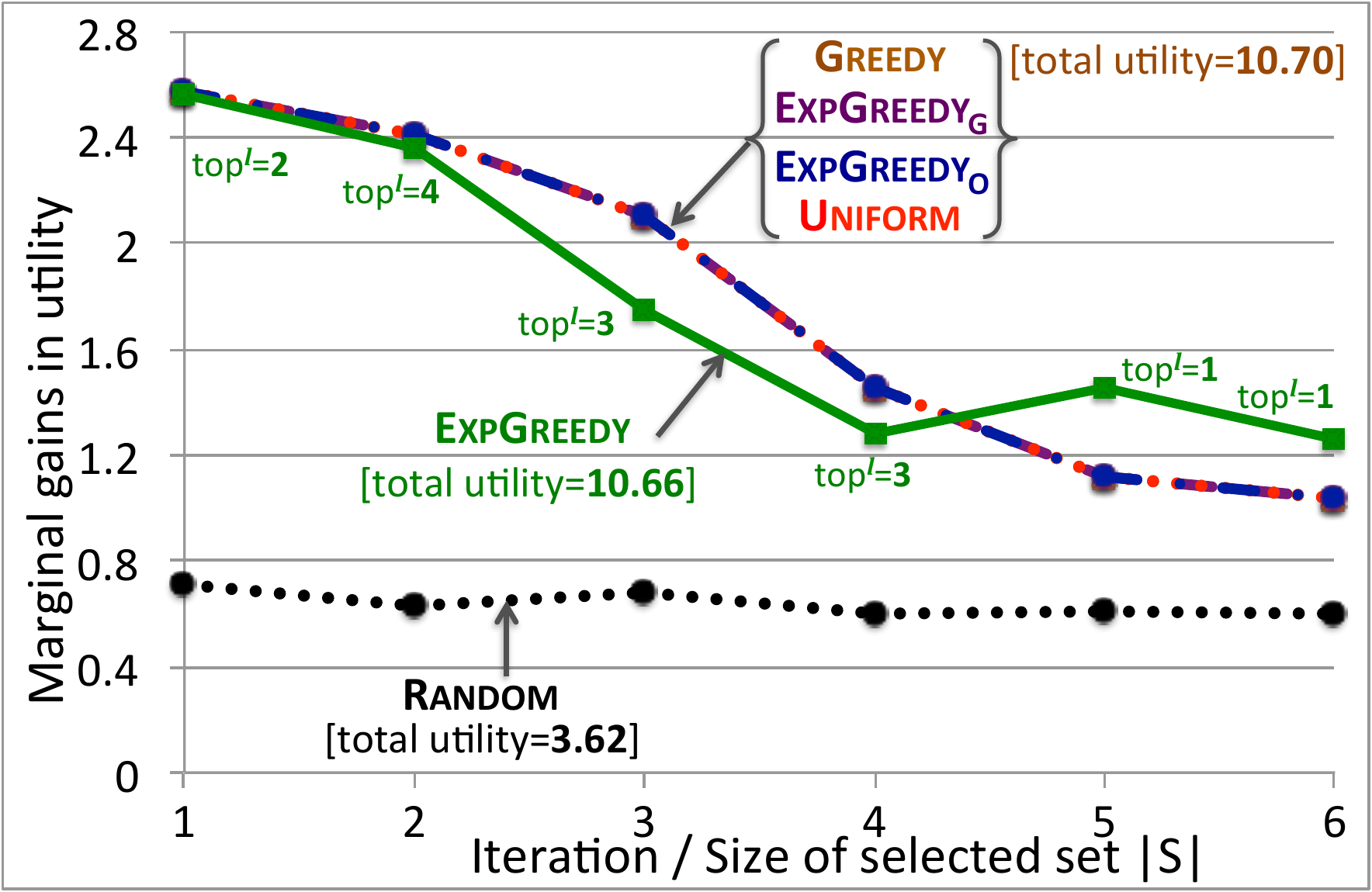}
    \label{fig.results.marginalgain}
   }
   \subfigure[Varying $\sigma^2$ for value queries]{
    \includegraphics[width=0.29\textwidth]{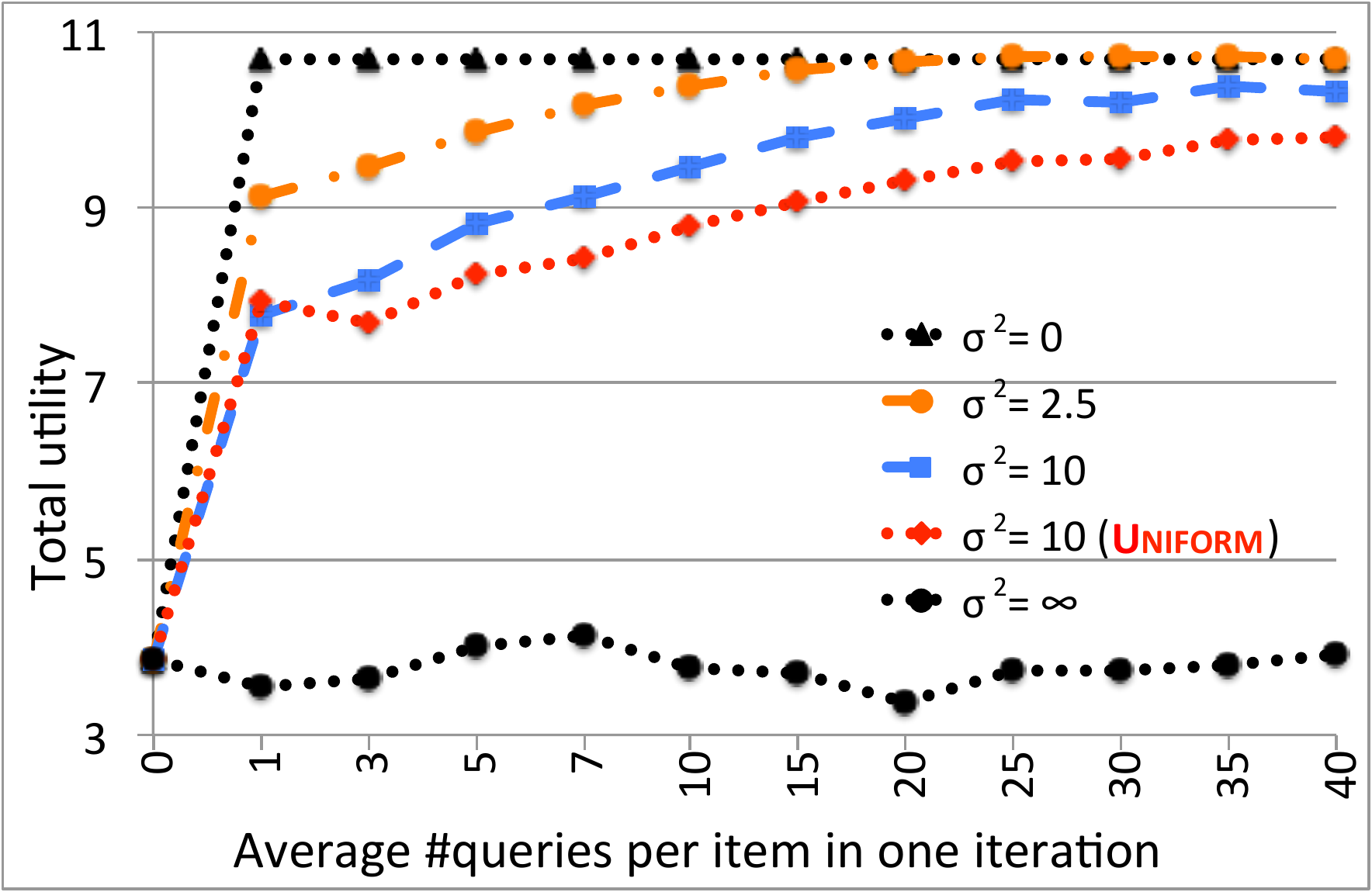}
     \label{fig.results.vary-sigma}
   }   
   \subfigure[Varying $\beta$ for preference queries]{
    \includegraphics[width=0.29\textwidth]{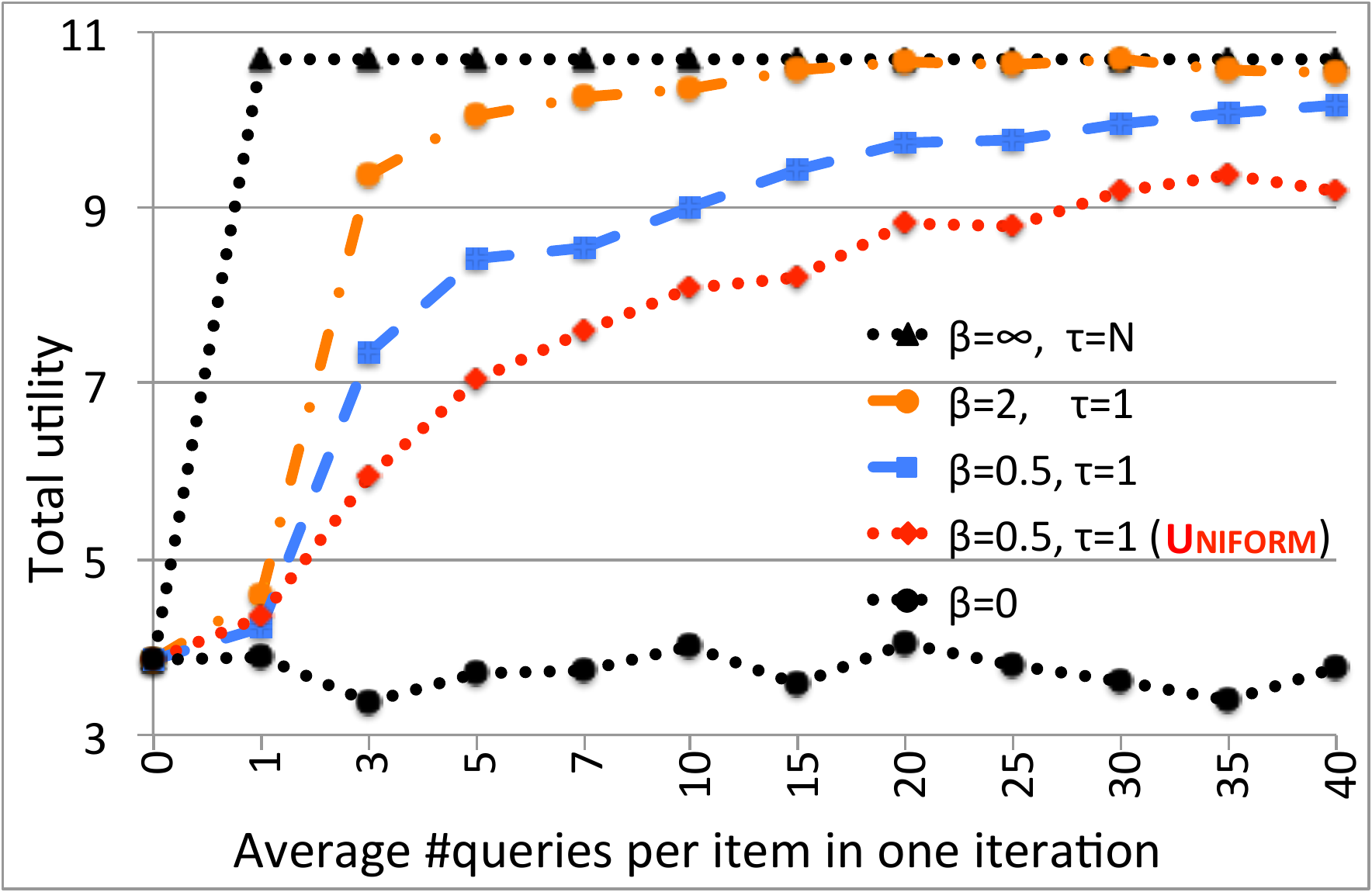}
     \label{fig.results.vary-beta}
   }   
   \subfigure[Varying $\tau$ for preference queries]{
    \includegraphics[width=0.29\textwidth]{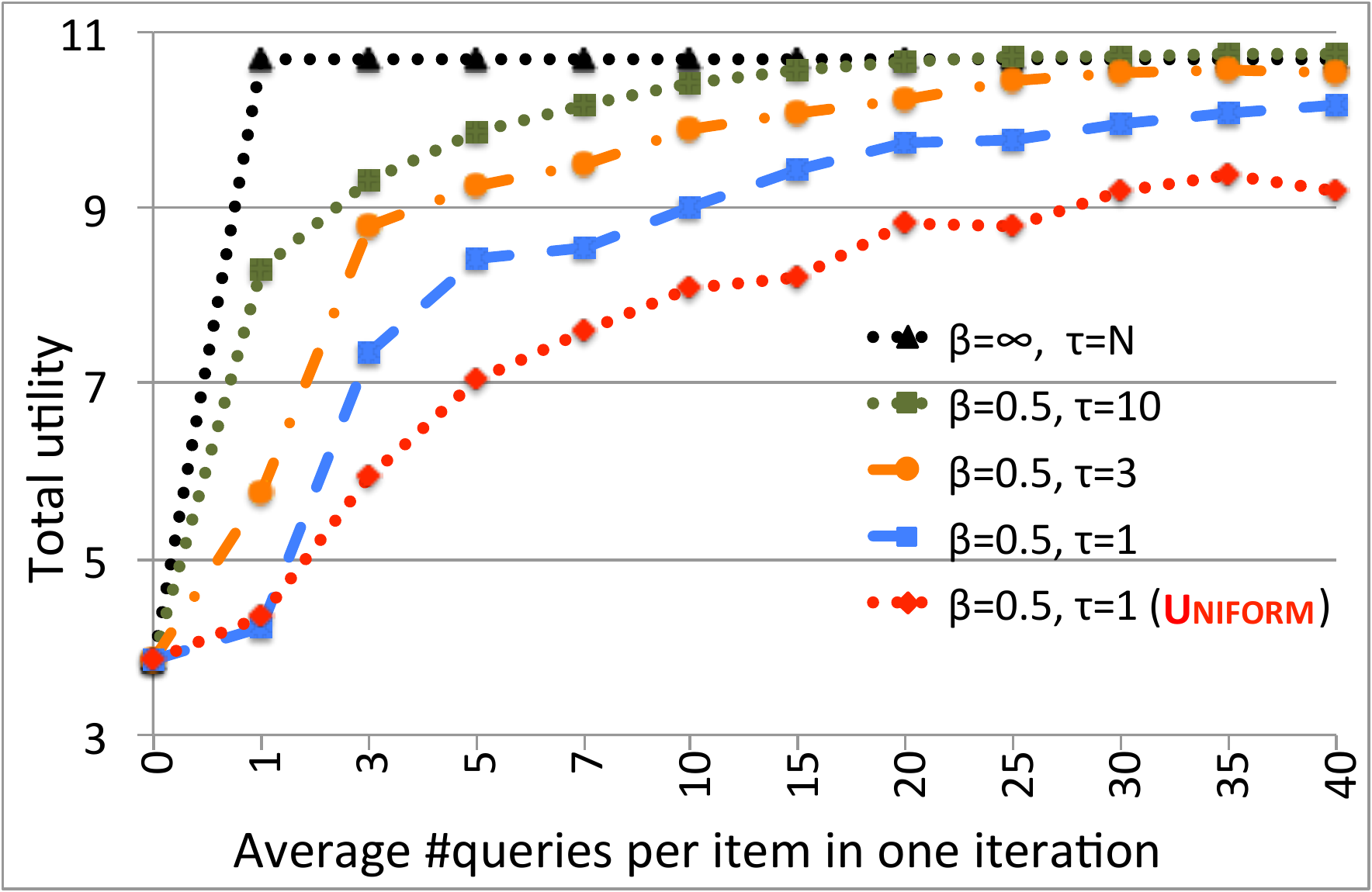}
     \label{fig.results.vary-tau}
   }   
\caption{\footnotesize Experimental results using synthetic function $f$ and simulated query responses: (a)--(d) results are for value queries and (e)--(f) results are for preference queries. (a) \submM dramatically reduces the sample complexity compared to \uniform and the other adaptive baselines.  (b) \submM adaptively allocates queries to identify largest gap $\Delta_l$. (c) An execution instance showing the marginal gains and sizes of $top^l$ solutions returned by \blbox{}. (d)--(f) \submM is robust to noise and outperforms \uniform baseline. }
\label{fig.results.syn}
\end{figure*}

\subsection{Experimental Setup}
{\bf Utility function $f$ and set of items $\actionset$.} 
\looseness -1 In the synthetic experiments we assume that there is an underlying submodular utility function $f$ which we aim to maximize. 
The exploration module \blbox{} performs value or preference queries, and receives noisy responses based on model parameters and the marginal gains of the items for this function. 
For our experiments, we constructed a realistic probabilistic coverage utility function following the ideas from \citet{2009_blogosphere} 
over a ground set of $N=60$ items.
Details of this construction are not important for the results stated below and can be found in Appendix~E of the extended version of this paper~\cite{singla16noisy-longer}.


{\bf Benchmarks and Metrics.}
We compare several variants of our algorithm, referred to by $\submM_\theta$, where $\theta$ refers to the parameters used to invoke \blbox{}. In particular, $\theta=O$ means $k'=1$ and $\epsilon'=\sfrac{\epsilon}{\nrItems}$ (i.e., competing with $(1 - \frac{1}{e})$ of $f(S^{opt})$); $\theta=G$ means $k'=1$ but $\epsilon'=0$ (i.e., competing with $f(S^{greedy})$); omitting $\theta$ means $k'=\nrItems$ and $\epsilon=\sfrac{\epsilon}{\nrItems}$. As benchmarks, we compare the performance with the deterministic greedy algorithm \greedy (with access to noise-free evaluations), as well as with random selection \random. As a natural and competitive baseline, we compare our algorithms  against \uniform \cite{kempe2003maximizing,krause05near} --- replacing our \blbox module by a naive exploration module that uniformly samples all the items for best item identification.  For all the experiments, we used PAC parameters $(\epsilon=0.1, \delta=0.05)$ and a cardinality constraint of $\nrItems=6$.

\subsection{Results}

{\bf Sample Complexity.}
In Figure~\ref{fig.results.sample-complexity}, we consider value queries, and compare the number of queries performed by different algorithms under varying noise-levels until convergence to the solution with the desired guarantees. For variance $\sigma^2$, we generated the query responses by sampling uniformly from the interval $[\mu - \sigma^2, \mu + \sigma^2]$, where $\mu$ is the expected value of that query.
For reference, the query cost of \greedy with access to the unknown function $f$ is marked (which equals $N \cdot \nrItems$).  The sample complexity differs by orders of magnitude, i.e., the number of queries performed by \submM{} grows much slower than that of $\submM_G$ and $\submM_O$. The sample complexity of \uniform is worse by further orders of magnitude compared to any of the variants of our algorithm.

{\bf Varying $\sigma^2$ for value queries and ($\beta$, $\tau$) for preference queries.}
\looseness -1 Next, we investigate the quality of obtained solutions for a limited budget on the total number of queries that can be performed. 
Although convergence may be slow, one may still get good solutions early in many cases.
In Figures~\ref{fig.results.vary-sigma}--\ref{fig.results.vary-tau} we vary the available average budget per item per iteration on the x-axis --- the total budget available in terms of queries that can be performed is equivalent to $N \cdot \nrItems$ times the average budget shown on x-axis. In particular, we compare the quality of the solutions obtained by \submM for different $\sigma^2$ in Figure~\ref{fig.results.vary-sigma} and for different $(\beta,\tau)$ in Figure~\ref{fig.results.vary-beta}--\ref{fig.results.vary-tau}, averaged over 50 runs. The parameter $\beta$ controls the noise-level in responses to the preference queries, whereas  $\tau$ is the algorithm's parameter indicating how many pairwise comparisons are done in one query to reduce variance. For reference, we show the extreme case of $\sigma^2 = \infty$ and $\beta = 0$ which is equivalent to \random; and the case of $\sigma^2 = 0$ and $(\beta = \infty, \tau = N)$ which is equivalent to \greedy. In general, for higher $\sigma$ in value queries, and for lower $\beta$ or smaller $\tau$ in preference queries, more budget must be spent to achieve solutions with high utility.

{\bf Comparison with \uniform exploration.}
In Figure~\ref{fig.results.vary-sigma} and Figure~\ref{fig.results.vary-beta}--\ref{fig.results.vary-tau}, we also report the quality of solutions obtained by \uniform  for the case of $\sigma^2=10$ and $(\beta=0.5,\tau=1)$ respectively.  As we can see in the results, in order to achieve a desired value of total utility, \uniform may require up to 3 times more budget in comparison to that required by \submM.
Figure~\ref{fig.results.budget-allocation} compares how different algorithms allocate budget across items (i.e., the distribution of queries), in one of the iterations. We observe that \submM{} does more exploration across different items compared to $\submM_G$ and $\submM_O$. However, the exploration is heavily skewed in comparison to \uniform because of the adaptive sampling by \blbox{}.  
Figure~\ref{fig.results.marginalgain} shows the marginal gain in utility of the considered algorithms at different iterations for a particular execution instance (no averaging over multiple executions of the algorithms was performed). For \submM, the size of $top^l$ solutions returned by \blbox{} in every iteration is indicated in the Figure, demonstrating that \blbox{} adaptively allocates queries to efficiently identify the largest gap $\Delta_l$.
\section{Image Collection Summarization}\label{sec.experiments.crowd}
We  now present results on a crowdsourced image collection summarization application, performed on Amazon's Mechanical Turk platform. As our image set $\actionset$, we retrieved 60 images in some way related to the city of Venice from Flickr, cf., Figure~\ref{fig.results.collection}. A total of over 100 distinct workers participated per summarization task.

The workers were queried for pairwise preferences as follows. They were told that our goal is to summarize images from Venice, motivated by the application of selecting a small set of pictures to send to friends after returning from a trip to Venice. The detailed instructions can be found in Appendix~F of the extended version of this paper \cite{singla16noisy-longer}. We ran three instances of the algorithm for three distinct summarization tasks for the themes $(i)$ Venice,   $(ii)$ Venice Carnival,  and $(iii)$ Venice Cathedrals. The workers were told the particular theme for summarization. Additionally, the set of images already selected and two proposal images $a$ and $b$ were shown. Then they were asked which of the two images would improve the summary more if added to the already selected images.

For running \submM, we used an average budget of $25$ queries per item per iteration and $\tau = 3$, cf., Figure~\ref{fig.results.vary-tau}.  Note that we did not make use of the function $f$ constructed for the synthetic experiments at all, i.e., the function we maximized was not known to us. The results of this experiment are shown in Figure~\ref{fig.results.crowdsummary}, demonstrating that our methodology works in real-word settings, produces high quality summaries and captures the semantics of the task.

%% file: conclusion.tex
\section{Conclusions}\label{sec.conclusions}

We considered the problem of cardinality constrained submodular function maximization under noise, i.e., the function to be optimized can only be evaluated via noisy queries. We proposed algorithms based on novel adaptive sampling strategies to achieve high quality solutions with low sample complexity. Our theoretical analysis and experimental evaluation provide insights into the trade-offs between solution quality and sample complexity. Furthermore, we demonstrated the practical applicability of our approach on a crowdsourced image collection summarization application.



%% file: appendix.tex

\onecolumn
\appendix
{\allowdisplaybreaks

\input{appendix_theorem_proof.tex}

\input{appendix_greedy.tex}


\input{appendix_theorem_proof_sample_complexity.tex}

\input{appendix_borda_score.tex}

\input{appendix_experiments_image_summarization.tex}


}

%% file: appendix_theorem_proof.tex

\section{Appendix A.  Proof of Theorem 1}
\label{app.theorem_proof}

In the following, we provide the proof of Theorem~\ref{thm2}.

\begin{proof}
  \emph{This proof adopts the proof of~\cite[Theorem 3.1]{2014_submodular_maximization_with_cardinality_constraints} to our setting.}
  
  Let $OPT \in \arg \max_{B \subseteq \actionset, |B| \leq \nrItems} f(B)$.
  For $1 \leq i \leq \nrItems$, let $S_{i-1}$ be the set of items selected before $i$-th iteration, and $A$ be the set returned by \blbox{} in the $i$-th iteration. Furthermore, let $A' \subseteq OPT \setminus S_{i-1}$ be of size $|A|$.
  We have that with probability at least $1-\delta'$,
  \begin{align}
    \mathbb{E}[f(s|S_{i-1})] &= |A|^{-1} \sum_{a \in A} f(a|S_{i-1}) \\
    	&\stackrel{(i)}{\geq}  |A|^{-1} \sum_{a \in A'} f(a|S_{i-1}) - \epsilon' \\
	    &= |A|^{-1} \sum_{a \in OPT \setminus S_{i-1}} f(a|S_{i-1}) - \epsilon' \\
	    &\stackrel{(ii)}{\geq} |A|^{-1} (f(OPT \cup S_{i-1}) - f(S_{i-1})) - \epsilon' \\
	    &\stackrel{(iii)}{\geq} |A|^{-1} (f(OPT) - f(S_{i-1})) - \epsilon' \\
	    &\stackrel{(iv)}{\geq} \nrItems^{-1} (f(OPT) - f(S_{i-1})) - \epsilon',
  \end{align}
  where the expectation is over elements $s$ drawn uniformly at random from $A$, $(i)$ is by the properties of \blbox{},  $(ii)$ by submodularity, $(iii)$ by monotonicity of $f$, and $(iv)$ because $|A| \leq k' \leq \nrItems$.
  Thus,
   \begin{align}
    \mathbb{E}[f(s|S_{i-1})] &\geq  \nrItems^{-1} (f(OPT) - \mathbb{E}[f(S_{i-1})]) - \epsilon',
  \end{align}
  The rest is standard, i.e.\ rearranging terms yields
  \begin{align}
    f(OPT) - \mathbb{E}[f(S_i)] \leq \left( 1 - \frac{1}{\nrItems} \right) [f(OPT) - \mathbb{E}[f(S_{i-1})]  + \epsilon'.
  \end{align}
    Note that the above inequality holds with probability $1-\delta'$ for one of the iteration.
  Repeatedly applying the above recursion gives
  \begin{align}
    f(OPT) - \mathbb{E}[f(S_i)] &\leq \left( 1 - \frac{1}{\nrItems} \right)^i [f(OPT) - \mathbb{E}[f(S_{0})]  + \sum_{i'=0}^{i-1} \left( 1 - \frac{1}{\nrItems} \right)^{i'}  \epsilon' \\
        &\leq \left( 1 - \frac{1}{\nrItems} \right)^i [f(OPT) - \mathbb{E}[f(S_{0})]  + i \epsilon' \\
        &\leq \left( 1 - \frac{1}{\nrItems} \right)^i f(OPT)  + i \epsilon'.
  \end{align}  
  To this end, we get
  \begin{align}
    \mathbb{E}[f(S_\nrItems)] &\geq f(OPT) - \left( 1 - \frac{1}{\nrItems} \right)^\nrItems f(OPT)  - \nrItems \epsilon'\\
    	&= [1 - \left( 1 - \frac{1}{\nrItems} \right)^K] f(OPT)  - \epsilon \\
	    &\geq \left( 1 - \frac{1}{e} \right) f(OPT)  - \epsilon.
  \end{align} 
  
  By the union bound over $\nrItems$ iterations and using the fact that $\delta' = \frac{\delta}{\nrItems}$, the above equation holds with probability at least $1-\delta$. 
\end{proof}

%% file: appendix_greedy.tex

\section{Appendix B. Competing With \greedy}
\label{sec.exampleGreedy}

In this section, we present an example that illustrates why competing with the greedy algorithm's solution $S^{greedy}$ can be especially difficult  in a noisy setting. Consider a ground set $\actionset=\{a,b,c\}$ consisting of three items. For $\alpha > 0$, we define a submodular function $f$ on $2^\actionset$ as follows:
\begin{gather*}
  f(\emptyset) = 0 \\
  f(\{a\}) = 1, \quad f(\{b\}) = 1, \quad f(\{c\}) = 1 - 2\alpha \\
  f(\{a,b\}) = 2, \quad f(\{a,c\}) = 1.5 - \alpha, \quad f(\{b,c\}) = 1.5 - \alpha\\
  f(\{a,b,c\}) = 2
\end{gather*}

Assume now that we want to maximize $f$ by selecting a subset $S \subseteq \actionset$ of at most size 2. Clearly, in a noiseless setting, greedy will return $S^{greedy} = \{ a,b \}$ such that $f(S^{greedy}) = 2$. However, any algorithm picking item $c$ first will only observe a utility of at most $1.5 - \alpha$, i.e.\ $f(\{a,c\}) = 1.5 - \alpha$ and $f(\{b,c\}) = 1.5 - \alpha$.

In maximization under noise, ensuring that item $c$ is \emph{not} picked first requires that our algorithms are highly confident about the marginal gains of the items. More precisely, the confidence intervals for the maginal values of the items should be at least smaller than $2\alpha$.  To this end, consider the sample complexity bounds of our algorithms for competing with greedy, i.e.\ for $\epsilon' = 0$ and $k' = 1$:
\begin{align}
  T = \mathcal{O}\left(\frac{1}{\alpha^{2}} \log (\frac{1}{\delta' \alpha^{2}} ) \right),
\end{align}
Thus, when $\alpha \rightarrow 0$, the sample complexity goes to $\infty$.


%% file: appendix_theorem_proof_sample_complexity.tex

\section{Appendix C.  Proof Sketch of Theorem~\ref{thm.sample.complexity}}
\label{app.theorem_proof_sample_complexity}

For proving Theorem~\ref{thm.sample.complexity}, consider a simpler variant of Algorithm~\ref{alg.modmax}. Assume that we solve the $top^l$ identification problems for $l \in \{1, \ldots, k' \}$ independently (i.e., every instance has its own copy of the variables $v_i$ and $T_{t}(i)$) in parallel. Then, the sample complexity of each of these problems can be bounded according to~\cite[Theorem 5 in Appendix B]{chen2014combinatorial} and is essentially characterized by the gap $\Delta_{l}$. Our algorithm terminates whenever the termination condition of one of these $top^l$ problems is met -- this allows our algorithm to be adaptive to the largest $\Delta_l$, for $l\in \{1,\dots,k'\}$. I.e., as long as there is {\em some} value of $l$ with large $\Delta_l$, we will be able to enjoy low sample complexity. To conclude the proof, note that sharing variables between the instances only speeds up convergence of each $top^l$ problem.


%% file: appendix_borda_score.tex

\section{Appendix D. \blbox{} with Preference Queries --- Adjusting the $\epsilon'$ parameter} 
\label{app.borda_reduction}

As a rather subtle point, note that the termination condition in Step~$10$ of Algorithm~\ref{alg.modmax} ensures an $\epsilon'$ approximate solution in terms of the values used by the algorithm (i.e., Borda scores for the preference query model). However, the desired property of the \blbox in \eqref{cond.subsetop} requires  $\epsilon'$ error slack in terms of the original marginal gains of the items $f(a|S)$.  
Hence, a transformation of $\epsilon' \rightarrow \epsilon'_{borda}$ is needed, to ensure that  $\epsilon'_{borda}$ error slack in  Step~$10$ of Algorithm~\ref{alg.modmax} translates to $\epsilon'$ in \eqref{cond.subsetop}. In the Bradley-Terry-Luce preference model, this mapping can be obtained as  $\epsilon'_{borda} = \frac{\beta \cdot \epsilon'}{2 \cdot (N-1)}$. When the algorithm has no knowledge of the underlying model and, consequently, of how values are mapped to preference probabilities, $\epsilon'_{borda}$  is set to $0$.

%% file: appendix_experiments_image_summarization.tex

\section{Appendix E. Expertimental Setup for Synthetic Experiments --- Details} 
\label{app.image_summ}

The basic idea behind the synthetic experiments is that there is an underlying submodular utility function $f$ which we aim to maximize. The exploration module \blbox{} that we construct performs value or preference queries. We then simulate the noisy responses for \blbox{} based on model parameters and the marginal gains of the items for this function. In principle, we can use any submodular function (e.g., a concave function w.r.t the cardinality of the selected set). For our synthetic experiments, we considered the task of image collection summarization, keeping it in spirit of our real-world crowdsouring experiments. Given a set of images $\actionset$, the task is to select a subset of these images that best cover some theme of interest.  

\subsection{Utility function $f$}
For synthetic image collection summarization experiments, we constructed the submodular utility function $f(S)$ which we aim to maximize. We retrieved 350 images from Flickr along with their associated meta-data (asosciated tags, the number of tags in total, and the view-count), representing images in some way related to the city of Venice \footnote{We used the \texttt{flickr.photos\_search} function of Flickr API with tag query \emph{`Venice'}}.  Following~\citet{2009_blogosphere}, we constructed $f(S) = \sum_{j=1}^{20} w_j \, f_j(S)$, where each $f_j(S)$ is a probabilistic coverage function with range $[0,1]$ quantifying how well some topic $j$ is covered, and where $w_j$ is a weight quantifying the importance of topic $j$. 

More specifically, we did this as follows: For each of the 350 images from Flickr we are given the asosciated tags, the number of tags in total, and the view-count. As preprocessing, we kept only the 20 most frequent tags. For image $s$, this information can be represented by the triplet $(T_s, n_s, v_s)$, where $T_s$ is the set of present topics (equivalent to tags in our setting), $n_s = |T_s|$, and $v_s$ is the view-count. Without loss of generality, we can assume that each of the tags is identified by an integer in $\{1, \ldots, 20\}$. Using these tags, we define the utility function $f(S) = \sum_{j=1}^{20} w_j \, f_j(S)$, where each $f_j(S)$ is a set function with range $[0,1]$ quantifying how well topic $j$ (this corresponds to the $j$th tag) is covered, and where $w_j$ is a weight quantifying the importance of topic $j$. These weights are selected as $w_j = 20 \tfrac{\sum_a \mathbf{1} [j \in T_a]}{\sum_a n_a}$ such that  $\sum_{j=1}^{20} {w_j} = 20$ (normalization to any other value is clearly possible, resulting in a different scaling of the observed function values).\footnote{We obtained the following topics and corresponding weights: \texttt{gondola	1.91; water	1.91; canal	1.47; bridge 1.32; art	1.18; rialto	1.18; ponte	1.03; piazzasanmarco	0.88; bw	0.88; street	0.88; night	0.74; murano	0.74; blackwhite	0.74; boat	0.74; blue	0.74; grandcanal	0.74; girl	0.74; carnival	0.74; architecture	0.74; sunset	0.74}}.
 
\subsection{Set of items $\actionset$} 
In order to construct our ground set $\actionset$, we downloaded another set of 60 test images (along with the associated meta-data) that formed our test collection that we seek to summarize \footnote{30 test images were retrieved for the tag query \emph{`Venice'} and remaining 30 for the query \emph{`Venice \& Church'}. We ensured that these 60 images were from different users than the 350  images used to build model}. These test images are shown in Figure~\ref{fig.results.collection}. Our goal is to summarize this test collection in simulations for the theme \emph{Venice} via the function constructed above. 

To this end, we define the coverage of a set of images $S$ for topic $j$ as $f_j(S) = 1 - \prod_{s \in S} (1-f'_j(s))$, i.e., as a type of probabilistic coverage~\cite{2009_blogosphere}, where $f'_j(s) \propto \log(v_s) \tfrac{\mathbf{1}[j \in T_s]}{n_s}$ normalized to $[0,1]$ across all $j$ and $s$. Clearly, more advanced models could be used for scoring image collection summaries, e.g., V-ROUGE which is a scoring function tailored for the task of image collection summarization~\cite{2014_image_summarization}.